%% file: main.tex
\documentclass{article}

\usepackage[final]{NFFL2021}

\usepackage[utf8]{inputenc} 
\usepackage[T1]{fontenc}    
\usepackage[colorlinks=true,citecolor=blue]{hyperref}       
\usepackage{url}            
\usepackage{booktabs}       
\usepackage{amsfonts}       
\usepackage{amsmath}
\usepackage[noabbrev,capitalize]{cleveref}
\usepackage{nicefrac}       
\usepackage{microtype}      
\usepackage{xcolor}         
\usepackage{graphicx}
\usepackage{enumitem}
\usepackage{xspace}
\usepackage{caption}
\usepackage{tabularx}
\usepackage{multirow}
\usepackage{xcolor}
\usepackage{array}
\usepackage{amsthm}

\newtheorem{proposition}{Proposition}
\newtheorem{definition}{Definition}

\newcolumntype{P}[1]{>{\centering\arraybackslash}p{#1}}

\newcommand{\fedavg}{\textsc{FedAvg}\xspace}
\newcommand{\fedsgd}{\textsc{FedSGD}\xspace}
\newcommand{\ie}{\textit{i.e.,}\xspace}
\newcommand{\eg}{\textit{e.g.,}\xspace}

\title{Learning Federated Representations and Recommendations with Limited Negatives}

\author{
Lin Ning \\
Google Research \\
linning@google.com \\
\And 
Karan Singhal \\
Google Research \\
karansinghal@google.com \\
\And Ellie X. Zhou \\
Google \\
ezhou@google.com \\
\And Sushant Prakash \\
Google Research \\
sush@google.com
}

\begin{document}

\maketitle

\begin{abstract}
Deep retrieval models are widely used for learning entity representations and recommendations. Federated learning provides a privacy-preserving way to train these models without requiring centralization of user data. However, federated deep retrieval models usually perform much worse than their centralized counterparts due to non-IID (independent and identically distributed) training data on clients, an intrinsic property of federated learning that limits negatives available for training. We demonstrate that this issue is distinct from the commonly studied client drift problem. This work proposes \textit{batch-insensitive losses} as a way to alleviate the non-IID negatives issue for federated movie recommendations. We explore a variety of techniques and identify that batch-insensitive losses can effectively improve the performance of federated deep retrieval models, increasing the relative recall of the federated model by up to 93.15\% and reducing the relative gap in recall between it and a centralized model from 27.22\% - 43.14\% to 0.53\% - 2.42\%. We also open-source our code framework to accelerate further research and applications of federated deep retrieval models.
  
\end{abstract}

\input{Introduction} 
\input{FederatedDeepRetrivalModel} 
\input{BatchInsensitiveLosses}

\input{Evaluation}
\input{OpenSourceFramework} 
\input{Conclusion} 
\input{Acknowledgement}



\bibliographystyle{plain}
\bibliography{reference}

\clearpage
\appendix

\input{Appendix}

\end{document}

%% file: Introduction.tex
\section{Introduction} 

Recent years have witnessed the successes of deep retrieval models in many large-scale recommendation systems \cite{paul2016RecSys, Ma2018KDD, Yi2019RecSys, Yang2020WWW, Jiang2020WWW, Volkovs2017NIPS} and natural language tasks \cite{Henderson2017ArXiv, Gillick2018ArXiv, Yanga2018ACL,chidambaram2018learning}. While these models largely improve user experience by enabling personalization and representation learning, they can also raise privacy concerns as user data typically needs to be sent to a centralized server for training.

Federated learning (FL) is a decentralized training strategy in which clients collaborate with a coordinating server to train a machine learning model \citep{Brendan2017AISTATS}. It provides opportunities to leverage distributed client data to learn useful models while preserving user privacy. Bringing deep retrieval models to FL is a promising way to power recommendations and learn representations for users and items while addressing privacy concerns by reducing the centralization of user data.

In this work, we observe a challenge with training federated deep retrieval models: the insufficiency of negative examples available on a client’s device. A deep retrieval model (shown in \cref{fig:deep-retrieval-model}) learns embeddings representing users' contexts and items like movies, songs, or websites. For the model to learn meaningful embeddings, it generally uses two types of examples: positive examples and negative examples. The positive examples pull embeddings of the training \textit{(context, item)} pairs to be close together in an embedding space, while the negative examples push embeddings of unrelated pairs farther apart. Negative examples are typically required to prevent \emph{embedding space collapse}, where learned representations collapse to a single point and are no longer informative. 

Typically, negative examples are produced by sampling items from training data. In centralized training, training data is usually assumed to be independent and identically distributed (IID). Therefore, negatives can be sampled reliably from the overall training distribution. However, the IID data assumption does not hold for federated learning since clients generate data locally based on their circumstances. That means, on each device, negatives may not be present. Even if they are present, they may be relatively few and relatively similar. In practice, we observe that this leads to significant performance degradation when FL is applied naively, beyond the typical degradation observed for \eg classification in the FL setting.

This work focuses on understanding and alleviating the non-IID issue for deep retrieval models. We make the following \textbf{key contributions}:

\begin{itemize}[topsep=0pt,itemsep=-0.5ex,partopsep=0ex,parsep=1ex,leftmargin=5ex]
	\item Observe that naive federated training of deep retrieval models causes an unusually steep performance drop. Show that performance degradation is primarily caused by sampling negatives from non-IID data, not the typical client drift issue or other aspects of federated training.
	\item Introduce \textit{batch-insensitive losses} as a class of objectives to alleviate the issue in this setting.
	\item Perform empirical evaluation of different training objectives in a movie recommendation setting, showing that batch-insensitive losses enable performant federated deep retrieval.
	\item Release an open-source framework to accelerate further research and practical applications of federated deep retrieval models.\footnote{\href{https://git.io/federated_dual_encoder}{https://git.io/federated\_dual\_encoder}}
\end{itemize}

\textbf{Related Work: } Most previous research has focused on mitigating the non-IID data issue \cite{Hsieh2020ICML, Zhao2018arXiv, Li2018arXiv, Li2020ICLR, Li2021ICLR, Yu2020ICML, Yoon2021ICLR} for improving general model convergence in federated learning settings. These works study the general problem of heterogeneity in client data causing \emph{client drift} when clients perform multiple local computation steps, as in \fedavg \citep{karimireddy2020scaffold,karimireddy2020mime}. However, some of them require sharing some training data across clients \cite{Hsieh2020ICML, Zhao2018arXiv}, and most of them study image classification tasks. The developed techniques are not fully relevant to the deep retrieval model, which sees unusually severe performance degradation due to explicit reliance on sampling negatives for training. Though also caused by non-IID data, this issue is orthogonal to the client drift issue: it occurs even when clients perform one local update step. We can combine the techniques explored in this work with previous techniques for addressing client drift. \cite{Yu2020ICML} also studies the effectiveness of hinge loss and spreadout regularizer in federated learning. However, it focuses on classification tasks and only considers extreme cases where each user has no access to negative examples. Our work focuses on a more realistic setting, aiming to understand and increase the performance of representation learning and item recommendation with deep retrieval models when each client has access to some but limited and relatively similar negative examples. More broadly, we propose batch-insensitive losses, which can be generally valuable for alleviating non-IID data issues in other federated learning tasks.

%% file: FederatedDeepRetrivalModel.tex
\section{Federated Deep Retrieval Model}  
\label{sec:federated-deep-retrieval-model}
\begin{figure}[t]
\centering
    \includegraphics[width=0.45\columnwidth]{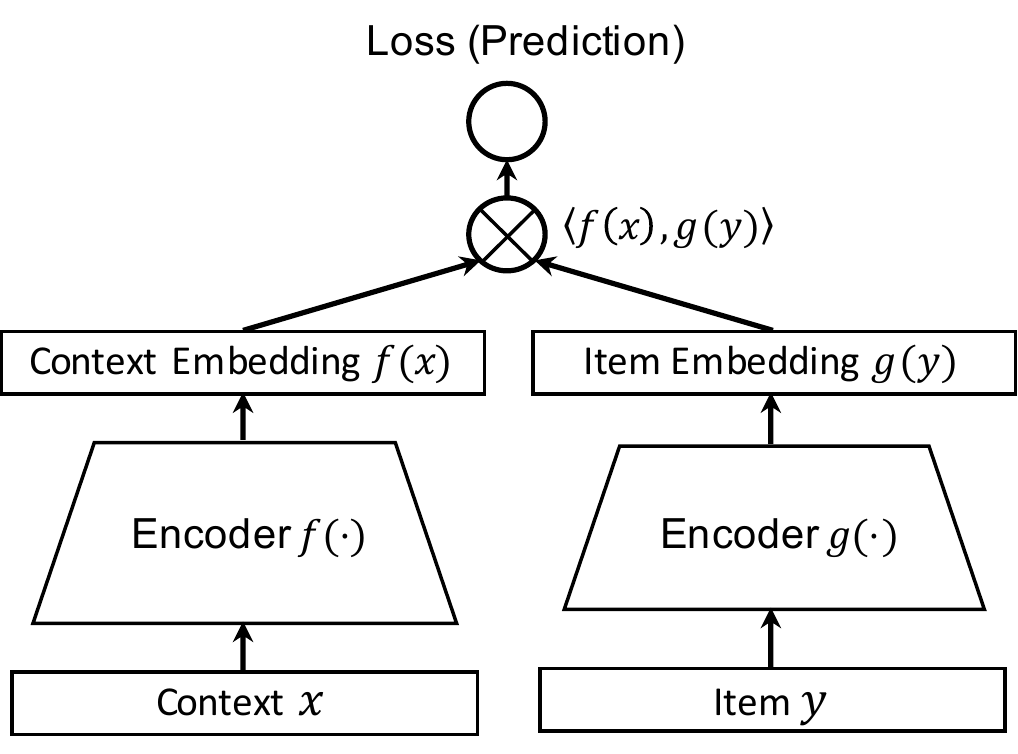}
    \caption{An illustration of a deep retrieval model.}
\vskip -0.2in
\label{fig:deep-retrieval-model}
\end{figure}

Deep retrieval models are also referred to as dual encoder, two-tower, or encoder-encoder models depending on the setting \cite{Gillick2018ArXiv, Yanga2018ACL, chidambaram2018learning,sountsov2016length,gillick2019learning}. This general framework has been successfully deployed in a variety of real-world applications in embedding and recommendation learning. As illustrated in \cref{fig:deep-retrieval-model}, a deep retrieval model consists of two encoders, each of which can be a fully-connected network, convolutional neural network, Transformer, and so on, depending on the task. The input consists of \textit{(context, item)} pairs encoded by the left and right encoders, respectively. For example, in a movie recommendation use-case, a context might be a sequence of previous movies a user has watched, and the item is the next movie they watch. The goals of applying a deep retrieval model in this setting would be to learn a model that can predict the next movie given an unseen sequence of previous movies and learn encoders that produce embeddings for users and movies. 
 
More formally, we denote the feature vectors representing contexts and items as $\mathbf{x}$ and $\mathbf{y}$. The two encoders are denoted as two parameterized functions $f(\cdot)$ and $g(\cdot)$, with $f(\mathbf{x})$ and $g(\mathbf{y})$ mapping $\mathbf{x}$ and $\mathbf{y}$ to a shared embedding space. The model outputs the similarity score between the encoded context and item, \eg the inner product of context and item embeddings, $s(\mathbf{x}, \mathbf{y}) = \big \langle f(\mathbf{x}), g(\mathbf{y}) \big \rangle$. A loss function is applied to enforce that positive examples (\ie similar context and item pairs) have high similarity, and negative examples have low similarity. 
Once the parameterized functions $f(\cdot)$ and $g(\cdot)$ are learned, the model can predict relevant items given a new context. The representations produced by $f(\cdot)$ and $g(\cdot)$ are general representations of user contexts and items and can also be used for other downstream applications, such as classification \cite{gillick2019learning,chidambaram2018learning}.

 \textbf{Loss Function: } The loss function plays a key role in training a deep retrieval model. The most commonly used one \cite{gillick2019learning,Yang2020WWW,Yi2019RecSys} is softmax cross-entropy loss over similarities :
$\ell (\mathbf{X}_i, \mathbf{Y}_i) = -\log (e^{s(\mathbf{X}_i, \mathbf{Y}_i)} / \sum_{\mathbf{Y}_j \in \mathcal{N}} e^{s(\mathbf{X}_i, \mathbf{Y}_j)})$. $\mathbf{X}$ and $\mathbf{Y}$ represent all the contexts and items in a batch, and $\mathcal{N}$ is a set of negative labels used to construct negative example pairs $(\mathbf{X}_i, \mathbf{Y}_j)$. The model is incentivized to maximize the similarity between positive example pairs and minimize the similarity between negative example pairs. Note that if it only did the former, then the embedding space produced by $f(\cdot)$ and $g(\cdot)$ would collapse. Therefore, negative examples are important to learn a good model. A standard method for getting negatives is to use in-batch negatives \citep{Yi2019RecSys,Yang2020WWW}, which means given a training batch and any ($\mathbf{X}_i$, $\mathbf{Y}_i$) pair in the batch, all other items $\mathbf{Y}_{j, j \neq i}$ in the same batch are treated as negatives for $\mathbf{X}_i$. We refer to this as \emph{batch softmax} below.

Federated training of a deep retrieval model involves three main steps in each training round. First, a central server sends the current model to several randomly sampled clients. Second, each sampled client trains on its dataset and updates its model locally. Finally, the local model updates are sent back to the server and aggregated to update the server model. The de-facto standard federated optimization method is \fedavg \citep{Brendan2017AISTATS}. Each client updates its model multiple times before sending the model update back to the server to be averaged. We also later refer to \fedsgd, where each client only runs a single local training step at each round, similar to standard distributed training. 

As discussed in \cref{sec:effect-of-non-iid-data}, naively applying federated learning to this setting produces a steep performance drop, worse than typical when comparing centralized and federated performance. We will show that training degrades due to non-IID negatives. Note that this is distinct from the \textit{client drift} phenomenon discussed in other works \citep{karimireddy2020scaffold}, which causes slight performance deterioration when clients take multiple local steps. In contrast, the problem we observe occurs even in the \fedsgd regime (see \cref{fig:centralized-tff-fedsgd}). This work aims to characterize this problem better and propose methods that enable federated deep retrieval models to perform comparably to centralized counterparts.

%% file: BatchInsensitiveLosses.tex
\section{Batch-Insensitive Losses}
\label{sec:batch-insensitive-loss}


We propose {\em batch-insensitive losses} as a potential solution to address the non-IID data issue. 

\begin{definition}[Batch-Insensitive Loss]
\label{def:batch-insensitive-loss}
Given a batch of N examples and two parameterized functions $f(\cdot)$ and $g(\cdot)$, with all contexts and items in the batch denoted as $\mathbf{X}$ and $\mathbf{Y}$, a loss function is batch-insensitive if it satisfies
\begin{equation}   
\label{eq:batch-insensitive-loss}
    \ell_{BI}(\mathbf{X}, \mathbf{Y}) = \frac{1}{N} \sum_{i = 0}^N \ell_{BI} (f(\mathbf{X}_i), g(\mathbf{Y}_i)).
\end{equation}
\end{definition}
It follows that applying a batch-insensitive loss over several batches of data in parallel (not in sequence) produces the same average loss and gradient update no matter how the examples are batched. We use this to show that we can produce the same gradient update between federated and centralized learning, providing a natural justification for batch-insensitive losses.
\begin{proposition} 
\label{prop:fedsgd-qual-sgd}
Let $\mathcal{C}$ be a collection of clients sampled at round $k$ for federated learning. Denote the aggregated update to the model for that round under \fedsgd as $\Delta_{k,fedsgd}(\mathcal{C})$. Let $\mathcal{E}$ be the collection of all training examples from clients in $\mathcal{C}$. Denote the model update of centralized training using SGD with all examples in $\mathcal{E}$ in a batch is $\Delta_{k,sgd}(\mathcal{E})$.
Using the same batch-insensitive loss $\ell_{BI}$ as training objective, the same model initialization $\Theta$, and the same learning rate $\eta$, we have
\begin{equation}
\Delta_{k,fedsgd} (\mathcal{C} | \ell_{BI}, \Theta) \equiv \Delta_{k,sgd} (\mathcal{E} | \ell_{BI}, \Theta)
\end{equation}
\end{proposition}
See \cref{appendix:fedsgd-equal-sgd} for proof. \cref{prop:fedsgd-qual-sgd} shows that \fedsgd with $\ell_{BI}$ approximates standard large-batch SGD when the number of clients per round is large (so $\mathcal{E}$ is representative of the centralized data). This motivates using batch-insensitive losses to mitigate the non-IID data issue.

Note that batch softmax loss does not have these properties and is batch-sensitive. We now describe specific batch-insensitive losses for the deep retrieval setting.

\subsection{Hinge Loss + Spreadout Regularizer}
\label{sec:hinge-spreadout-loss}

This loss is a variation of contrastive loss, composed of a positive term pushing positive examples together (hinge loss) and a negative term preventing embedding collapse (spreadout regularization). This combination was first introduced in \cite{Yu2020ICML}.

\textbf{Hinge Loss: }
Given a positive example pair ($\mathbf{x}$,$\mathbf{y}$), where $\mathbf{x}$ and $\mathbf{y}$ are context and item features, the hinge loss is defined as 
$
    \ell(\mathbf{x}, \mathbf{y}) = \max({0, \beta - f(\mathbf{x}) \cdot g(\mathbf{y})})^2,
$
where $f(\mathbf{x})$ is the context embedding, $g(\mathbf{y})$ is the item embedding, and $\beta$ is a tunable margin set to $0.9$ in this work. 

\textbf{Spreadout Regularization: }
Spreadout regularizer \cite{zhang2017learning} maximizes the spread of embeddings in an embedding space. Given an embedding vocabulary $V$ and the corresponding embedding weights $W$, spreadout regularizer can be formulated as 
$
    \ell_{sr}(W) = \sum_{v\in V} \sum_{v' \neq v}(-d^2(w_{v}, w_{v'})),
$
where $d$ is a measure of distance, \eg Euclidean distance or negative dot product. When combined with L2 normalization, the objective pushes embeddings in $W$ apart on a hypersphere. It can be used with any loss function in the form of
$
    \ell_{(\cdot)\cdot sr} = \ell_{(\cdot)} + \alpha \ell_{sr},
$
where $\ell_{(\cdot)}$ is the original loss function, $\ell_{sr}$ is the spreadout regularizer, and $\alpha$ trades off the regularization term and the original loss. 


Unlike softmax cross-entropy, hinge loss only considers positive examples; the loss can be trivially minimized by collapsing embeddings into a single point. To avoid collapse, we apply spreadout regularizer to the model's shared embedding table described in \cref{sec:experiment-settings}, pushing items in the embedding vocabulary to have orthogonal embeddings. The resulting combined loss pushes positive pairs closer while pushing negatives apart, resulting in a full loss for deep retrieval. Both hinge loss and spreadout regularizer are batch-insensitive based on \cref{def:batch-insensitive-loss}. It is easy to show that a linear combination of them is also batch-insensitive. 

\subsection{Global Softmax}
\label{sec:negative-sampling}

The \emph{global softmax} loss is defined as
$
    \ell (\mathbf{X}_i, \mathbf{Y}_i) = -\log (e^{f(\mathbf{X}_i) \cdot g(\mathbf{Y}_i)} / \sum_{\mathbf{Y}_j \in V} e^{f(\mathbf{X}_i) \cdot g(\mathbf{Y}_j)}),
$
where $V$ is the vocabulary of possible items to predict. Global softmax is an extreme case of negative sampling. For any ($\mathbf{X}_i$, $\mathbf{Y}_i$) pair, all the items $\mathbf{Y}_{j, j \neq i} \in V$ are treated as negatives for $\mathbf{X}_i$, and are used to calculate the softmax loss. For a larger scale model, only a subset of items can be sampled randomly instead. In either case, the loss is batch-insensitive on expectation according to \cref{def:batch-insensitive-loss}.

%% file: Evaluation.tex
\section{Evaluation} 
\label{sec:evaluation}
We evaluate the efficacy of batch-insensitive losses on the non-IID negatives issue with a movie recommendation task. We train and evaluate a deep retrieval model on the MovieLens 1M dataset \footnote{\href{https://grouplens.org/datasets/movielens/1m/}{https://grouplens.org/datasets/movielens/1m/}}. The model takes in a user's movie-watching history and predicts a relevant next movie for this user.

\subsection{Experiment Setting}
\label{sec:experiment-settings}
Below we describe details on the dataset, model, and tasks. Additional details can be found in our open-source code framework for federated deep retrieval (see \cref{sec:open-source-framework}).

\textbf{Dataset: } As shown in Table 1, the MovieLens 1M dataset contains approximately 1 million ratings from 6040 users on 3952 movies. Examples are created by taking moving "windows" of the movie sequence (sorted by timestamps) for each user, resulting in context inputs containing ten movie IDs and item inputs representing one next movie ID. For centralized training, examples are randomly shuffled across all users and split to train and test datasets. The train dataset has 894,752 examples, and the test dataset has 99,417 examples. We refer to these as {\em centralized datasets} later in this section. For federated training, all examples are grouped by user, forming a natural data partitioning across clients. The train and test examples are split by user ids, resulting in 4832 train, 603 validation, and 605 test users. We refer to these as {\em federated datasets}. We sample 100 clients for each training round.

\begin{figure}[t]
\centering
    \includegraphics[width=0.8\columnwidth]{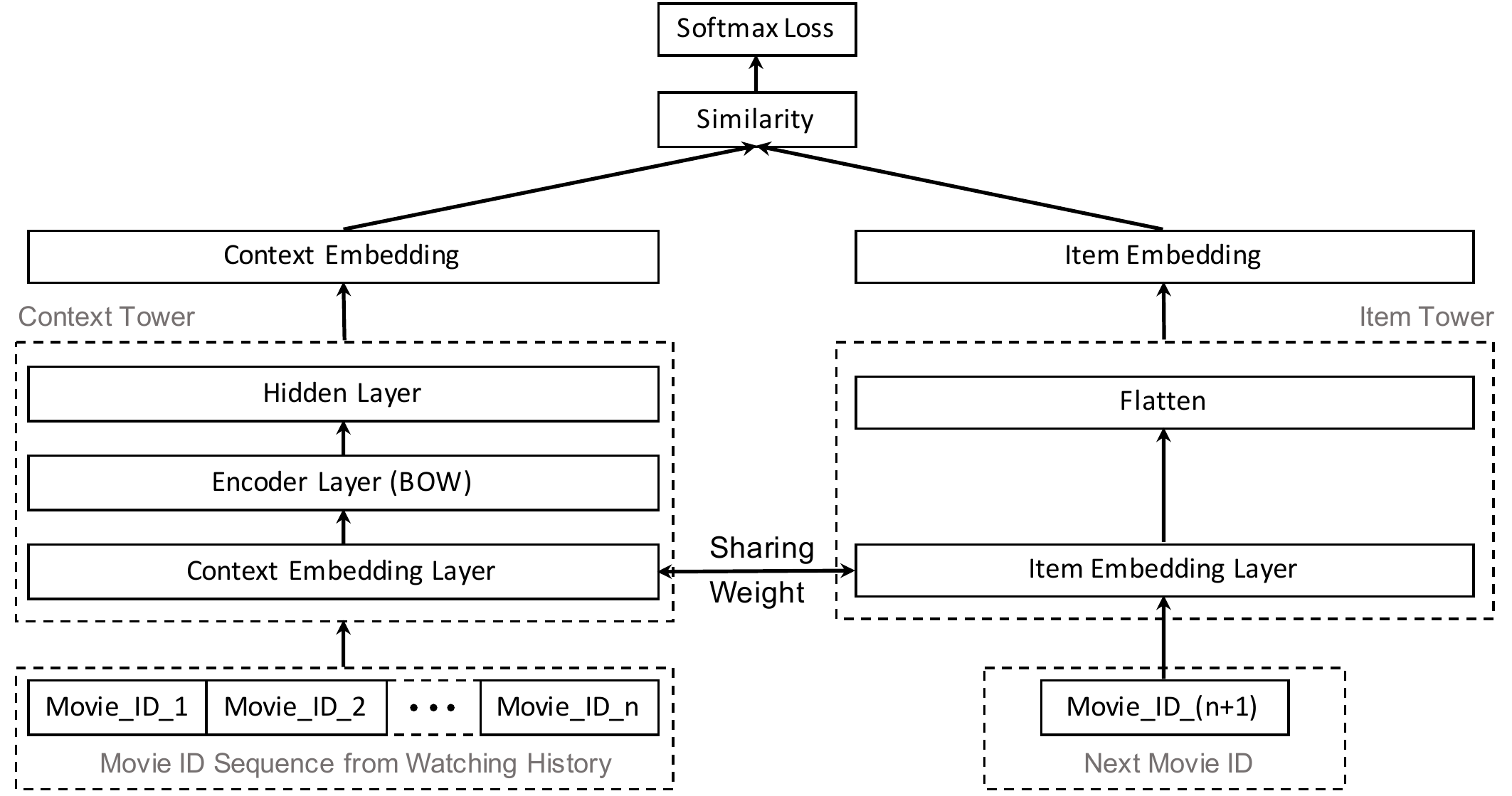}
    \vskip 0.1in
    \caption{ID-based deep retrieval model for movie recommendation.}
\label{fig:ID-based-deep-retrieval-model}
\vskip -0.2in
\end{figure}

\textbf{Model Architecture: }
\cref{fig:ID-based-deep-retrieval-model} illustrates the architecture of the ID-based deep retrieval model used for experiments. It takes a sequence of movie IDs (the movie watching history) as the context and the next movie ID as the item to form the \textit{(context, item)} pair. The context encoder is a bag-of-word encoder, and the item tower is a simple embedding lookup tower. The two towers share the same bottom embedding layer, which maps from movie ID to dense embedding. The two towers generate context and item embeddings respectively, and similarity is enforced between the positive (context, label) pairs. To make the comparison consistent and fair, we set the batch size to 16 for all experiments. The output dimension of the shared embedding layer is 16. The encoded context embedding and item embedding are also 16-dimensional and L2-normalized.

\textbf{Federated and Centralized Experiments: } One of our goals in this work is to reduce the gap between federated and centralized deep retrieval model performance. We run experiments for both centralized and federated training and compare performance. Interestingly, we observe that batch-insensitive losses can also improve centralized performance, so we also compare against this. We refer to these models as \emph{Improved Centralized} later in this section. In both settings, we measure test recall@k for $k \in [1, 5, 10]$, the fraction of examples for which the correct next movie is within the top k nearest item embeddings for an unseen context.

\subsection{Effect of Non-IID Data}
\label{sec:effect-of-non-iid-data}
\begin{figure}[t]
    \centering
    \begin{minipage}{0.45\textwidth}
        \centering
        \includegraphics[width=\textwidth]{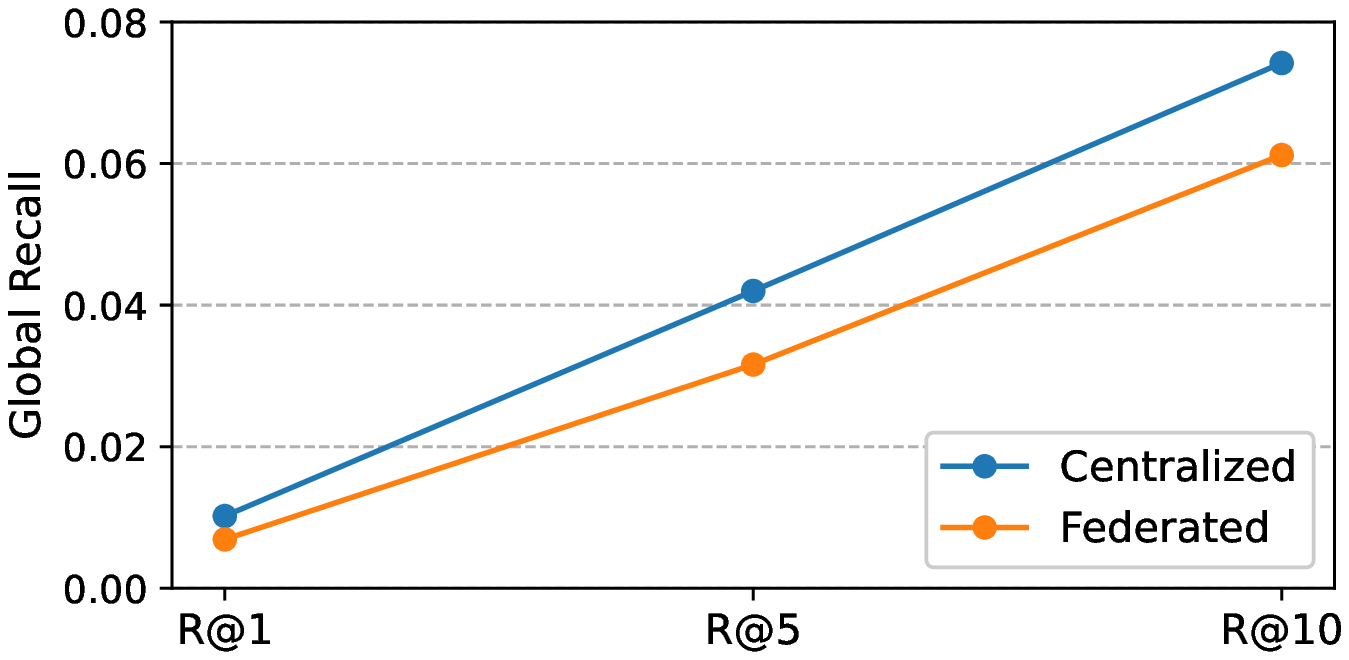} 
        \captionof{figure}{Model performance with batch softmax loss. Federated training uses \fedsgd.} 
        \label{fig:centralized-tff-fedsgd}
    \end{minipage}\hfill
    \begin{minipage}{0.45\textwidth}
        \centering
        \includegraphics[width=\textwidth]{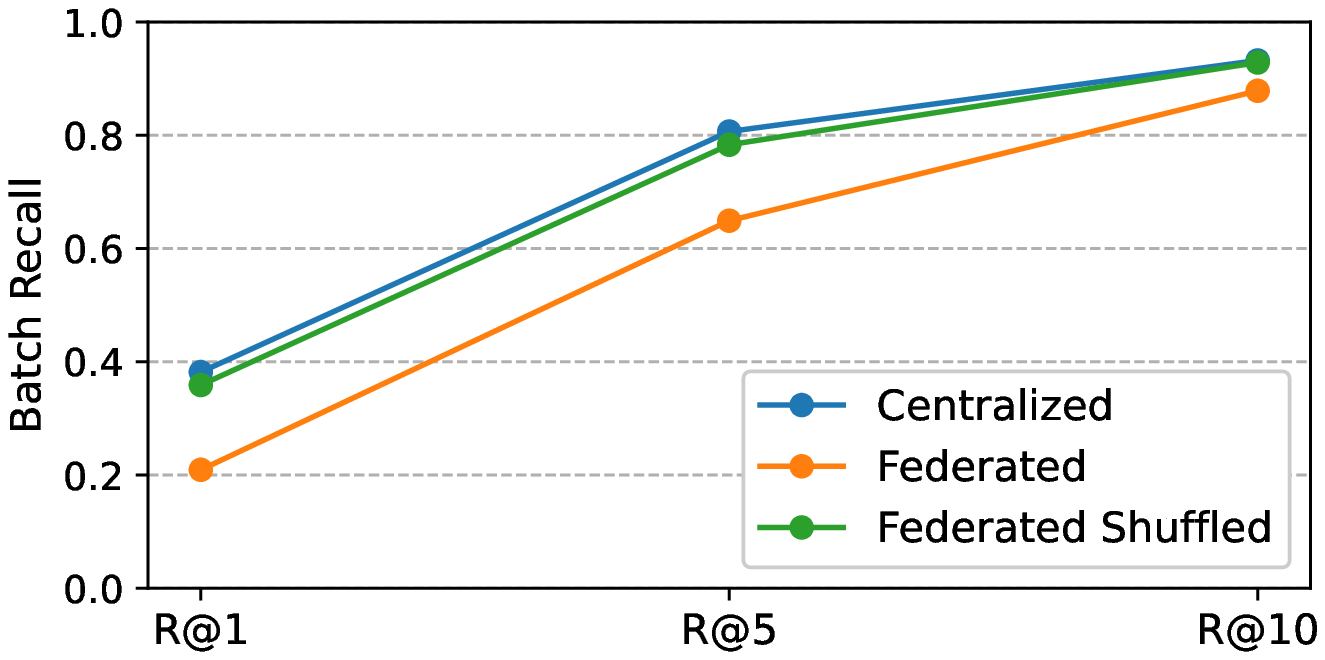} 
        \captionof{figure}{Model performance with batch softmax loss. Federated training uses \fedavg.} 
        \label{fig:non-IID-demnonstration}
    \end{minipage}
    \vskip -0.1in
\end{figure}

To study the effect of non-IID data on federated learning in this setting, we train the ID-based deep retrieval model on federated datasets using \fedavg \cite{Brendan2017AISTATS}. The model is trained with the standard batch softmax cross-entropy loss. We compare recall across items within a batch (batch recall) with centralized training. As shown in \cref{fig:non-IID-demnonstration}, the federated model experiences significant performance degradation, especially for recall@1. It is worth noting that performance degradation occurs even when training with \fedsgd, as illustrated in \cref{fig:centralized-tff-fedsgd}. It indicates that client drift, which occurs when clients take multiple local steps with non-IID data, is not the only cause.

To test whether non-IID data causes this performance degradation, we train a \emph{Federated Shuffled} model where data is IID across clients. In this experiment, all examples are shuffled across users while the number of examples per user remains the same. This ensures the same number of local steps taken on each client as before and isolates the effect of non-IID data on federated learning. As shown in \cref{fig:non-IID-demnonstration}, the Federated Shuffled result roughly matches the centralized result. We then conclude that non-IIDness causes the performance degradation, not federated training in itself. 

However, although shuffling examples across users could resolve the non-IID data issue, we cannot do it in practice due to privacy and communication limitations.

\begin{table}
\label{tab:dataset}
\caption{MovieLens 1M Dataset Statistics.  `E' is short for `Examples' and `U' is short for `Users'.}
\centering
\vskip 0.1in
\small
\begin{tabular}{lc}
\toprule
& Num \\
\midrule
Ratings & 1,000,209 \\
Users & 6040 \\
Movies & 3952 \\
\bottomrule
\end{tabular}
\quad
\begin{tabular}{lcc}
\toprule
& Centralized & Federated \\
\midrule
Split Strategy & By Example & By User \\
Train Data & 894,753 E & 4832 U \\
Test Data & 99,417 E & 605 U \\
\bottomrule
\end{tabular}
\vskip -0.1in
\end{table}

\subsection{Federated and Centralized Results}
This section studies the model performance with four different loss functions: batch softmax (BS), batch softmax with spreadout regularizer (BS+S), and two batch-insensitive losses (see \cref{sec:batch-insensitive-loss}): hinge loss with spreadout (H+S) and global softmax (GS). All the federated models are trained with \fedavg. We compare recall calculated globally across all items, which has no dependence on examples in a batch, enabling fair comparison.

\begin{figure}[t]
\centering
 \includegraphics[width=1\columnwidth]{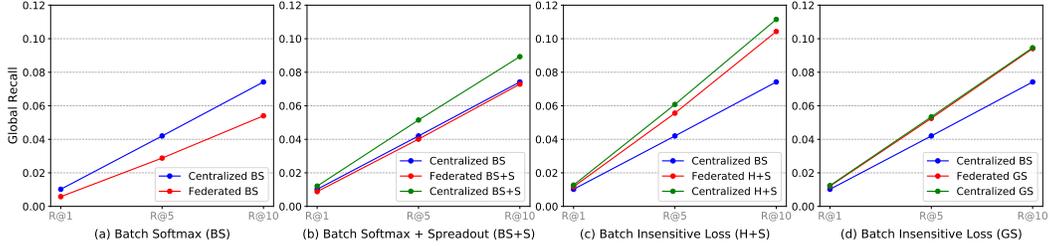}
\captionof{figure}{Comparison of centralized and federated models when training with different loss functions: batch softmax, batch softmax with spreadout regularizer, hinge loss with spreadout regularizer, and global softmax. The last two are batch-insensitive losses.}
\label{fig:4-methods-global-recall}
\end{figure}

\textbf{Batch Softmax (BS): } 
Batch softmax is the standard loss function for training a deep retrieval model. It is calculated with in-batch negatives as described in \cref{sec:federated-deep-retrieval-model}. \cref{fig:4-methods-global-recall}(a) shows a large gap between centralized and federated global recalls, similar to the batch recall results in \cref{sec:effect-of-non-iid-data}.

\textbf{Batch Softmax + Spreadout (BS+S): }
\cref{fig:4-methods-global-recall}(b) presents results of training with batch softmax combined with spreadout regularization. With spreadout regularizer, the recall values of  the federated model almost match those of the batch softmax centralized model, which is trained without spreadout regularizer. However, spreadout regularizer also improves centralized training. The federated model still performs significantly worse compared to the improved centralized model.

The results indicate that spreadout regularization itself is not enough to solve the issue with the non-IID negatives. Although spreadout regularizer pushes embeddings of unrelated pairs farther apart, batch softmax loss still depends on in-batch negatives and can still lead to worse model quality. Therefore, we need a loss function less affected by the training data distribution, motivating batch-insensitive losses.

\textbf{Batch-Insensitive Losses (H+S and GS): } 
\cref{fig:4-methods-global-recall}(c) and \cref{fig:4-methods-global-recall}(d) show the training results with the two types of batch-insensitive losses. With the combination of hinge loss and spreadout regularizer, both the improved centralized model and the federated model perform much better than the baseline model (\cref{fig:4-methods-global-recall}(d)). Also, the gap between improved centralized and federated models is much smaller than with batch softmax. With global softmax (\cref{fig:4-methods-global-recall}(c)), the federated model performs almost the same as the improved centralized model, and both perform significantly better than the batch softmax centralized model. Both of the results indicate that batch-insensitive loss alleviates the performance degradation caused by non-IID negatives effectively.

We caution that applying global softmax may not be appropriate in all settings. In these experiments, we use all the items in the movie vocabulary as the negatives. In practice, when dealing with items with large or unbounded vocabulary size, we may need other strategies as global softmax is computationally expensive.

\begin{table}[t]
\centering
\caption{An overall recall comparison between centralized, improved centralized, and federated models. The performance drop is calculated as $(R_c - R_{f}) / R_c $, where $R_c$ is the centralized global recall, and $R_f$ is the federated global recall.}
\vskip 0.1in
\label{tab:overall-comparison}
\scriptsize
\begin{tabular}{P{0.18in}P{0.42in}P{0.18in}P{0.18in}P{0.18in}p{0in}P{0.18in}P{0.18in}P{0.18in}P{0.28in}p{0in}P{0.23in}P{0.23in}P{0.23in}P{0.23in}}
\toprule
 & Centralized & \multicolumn{3}{c}{Improved Centralized} && \multicolumn{4}{c}{Federated} && \multicolumn{4}{c}{Performance Drop} \\

\cline{3-5} \cline{7-10} \cline{12-15}\noalign{\smallskip}
 & BS & BS+S & H+S & GS && BS & BS+S & H+S & GS && BS & BS+S & H+S & GS \\
 \midrule
R@1 & 1.02 & 1.21 & {\bf 1.27} & 1.24 && 0.58 & 0.88 & 1.17 & {\bf 1.21} && 43.14\% & 27.27\% & 7.87\% & {\bf 2.42}\% \\ 
R@5 & 4.2 & 5.15 & {\bf 6.08} & 5.34 && 2.88 & 4.01 & {\bf 5.56} & 5.25 && 31.43\% & 22.14\% & 8.55\% & {\bf 1.69}\% \\ 
R@10 & 7.42 & 8.93 & {\bf 11.15} & 9.46 && 5.4 & 7.29 & {\bf 10.43} & 9.41 && 27.22\% & 18.37\% & 6.46\% & {\bf 0.53}\% \\ 
\bottomrule
\end{tabular}
\end{table}

\textbf{Overall Comparison: }
\cref{tab:overall-comparison} gives an overall performance comparison between centralized, improved centralized, and federated models under different losses. 

Training with batch-insensitive losses (H+S, GS) achieves the highest recall for both centralized and federated models. In particular, hinge loss with spreadout regularizer appears to perform slightly better than global softmax in terms of absolute recall, but both perform significantly better than batch-sensitive losses (BS, BS+S).

We also observe that global softmax incurs the smallest performance gap between centralized and federated training. Hinge loss with spreadout regularizer has the next smallest performance gap, and batch-sensitive techniques have a more significant performance gap as expected. We expect that the remaining performance drop between centralized and federated models results from client drift (clients are still taking multiple local steps). This suggests that combining batch-insensitive losses with approaches to address client drift may be a promising future direction. 

%% file: OpenSourceFramework.tex
\section{Open-Source Framework}  
\label{sec:open-source-framework}

We are releasing a general code framework for experimenting with federated deep retrieval models built on the TensorFlow Federated library\footnote{\href{https://www.tensorflow.org/federated}{https://www.tensorflow.org/federated}}. The code is released under Apache License 2.0. The framework enables reproduction of our experiments and provides a flexible, well-documented interface for researchers to train federated and centralized deep retrieval models with different models and losses. We provide libraries for training and evaluation for MovieLens next movie prediction, which can be easily extended for new tasks. We hope that this framework spurs further research and lowers the barrier to more practical applications.

%% file: Conclusion.tex
\section{Conclusion} 

This work investigates the effect of non-IID negatives on federated training of deep retrieval models and proposes batch-insensitive losses to alleviate the issue. We compare model performance using various loss functions and show that batch-insensitive losses produce better federated deep retrieval models that can approximately match centralized models. We also open-source our code framework to accelerate future research and applications. Note that our proposed techniques do not directly address the separate, well-studied issue of client drift when clients do multiple steps of local training–approaches addressing this issue are complementary and can be combined with our work. 

%% file: Acknowledgement.tex
\section*{Acknowledgements}
We thank Warren Morningstar, Chung-Ching Chang, and Zachary Garrett and for their helpful comments and discussions. We also thank Warren Morningstar for his contribution to the federated training pipeline.

%% file: Appendix.tex
\section{Proof of Proposition 1}
\label{appendix:fedsgd-equal-sgd}

\begin{proof}
Assume $\mathcal{C}$ contains $M$ clients. A client $c_i \in \mathcal{C}$ has $N_i$ examples locally. With \fedsgd, a client $c_i$ only train model for a single step for each training round, with a batch size of $N_i$. Since the model is trained with a batch insensitive loss $\ell_{BI}$, with \cref{eq:batch-insensitive-loss}, we derive that the local model gradient of client $c_i$ at the end of training round $k$ is 
\begin{align}
\begin{split}
    \nabla \ell_{C_i} 
    &= \nabla \frac{1}{N_i} \sum^{N_i}_{j=0} \ell_{BI} (f(\mathbf{X}_j), g(\mathbf{Y}_j)) \\
    &= \frac{1}{N_i} \sum^{N-i}_{j=0} \nabla \ell_{BI} (f(\mathbf{X}_j), g(\mathbf{Y}_j))
\end{split}
\end{align}

The local model gradient of all the clients are aggregated to update the server model. Therefore, the server model update at step $k$ is

\begin{align}
\begin{split}
    \Delta_{k,fedsgd}(\mathcal{C}|\ell_{BI}, \Theta) 
    &= \eta_{s,fedsgd}  \cdot \frac{\sum^{M}_{i=0} \nabla \ell_{C_i}}{\sum^M_{i=0} N_i} \\
    &= \eta_{s,fedsgd}  \cdot \frac{\sum^M_{i=0}\sum^{N_i}_{j=0} \nabla \ell_{BI} (f(\mathbf{X}_j), g(\mathbf{Y}_j)) }{\sum^{M}_{i=0} N_i}    
\end{split}
\end{align}

Let $N_{\mathcal{E}}$ be the number of total examples in $\mathcal{E}$, we have $N_{\mathcal{E}} = \sum^M_{i=0} N_i$. Then the server model update becomes
\begin{equation}
\label{eq:model-update-fedsgd}
    \Delta_{k,fedsgd}(\mathcal{C}|\ell_{BI}, \Theta) 
    =\eta_{s,fedsgd}  \cdot \frac{1}{N_{\mathcal{E}}}\sum^{N_{\mathcal{E}}}_{i=0} \nabla \ell_{BI} (f(\mathbf{X}_i), g(\mathbf{Y}_i))
\end{equation}

For centralized training with SGD with all examples in $\mathcal{E}$ in a batch, the model update at step k is
\begin{align}
\label{eq:model-update-sgd}
\begin{split}
    \Delta_{k,sgd}(\mathcal{E}|\ell_{BI}, \Theta)
    &= \eta_{s,sgd} \cdot \nabla \frac{1}{N_{\mathcal{E}}} \sum^{N_{\mathcal{E}}}_{i=0} \ell_{BI} (f(\mathbf{X}_j), g(\mathbf{Y}_j))  \\
    &= \eta_{s,sgd} \cdot \frac{1}{N_E} \sum^{N_{\mathcal{E}}}_{i=0} \nabla \ell_{BI} (f(\mathbf{X}_j), g(\mathbf{Y}_j))
\end{split}
\end{align}

Note that  $\eta_{s, fedsgd} = \eta_{s, sgd}$. Therefore, with \cref{eq:model-update-fedsgd} and \cref{eq:model-update-sgd}, we prove that
\begin{equation*}
    \Delta_{k,fedsgd}(\mathcal{C}|\ell_{BI}, \Theta) \equiv \Delta_{k,sgd}(\mathcal{E}|\ell_{BI}, \Theta)
\end{equation*}
\end{proof}